\documentclass[sigplan,screen,preprint]{acmart}
\AtBeginDocument{%
  \providecommand\BibTeX{{%
    \normalfont B\kern-0.5em{\scshape i\kern-0.25em b}\kern-0.8em\TeX}}}

\usepackage{lipsum}  
\usepackage{subcaption}
\usepackage{bm}
\usepackage{algorithm2e}
\newcommand{\argmin}[1]{\underset{#1}{\operatorname{arg}\,\operatorname{min}}\;}
\setcopyright{none}
\copyrightyear{2022}
\acmYear{2022}
\acmDOI{XXXXXXX.XXXXXXX}

\begin{document}

\title{Closing the Gap between Client and Global Model Performance in Heterogeneous Federated Learning}

\author{Hongrui Shi}
\email{hshi21@sheffield.ac.uk}
\affiliation{%
  \institution{University of Sheffield}
  \country{}
}

\author{Valentin Radu}
\email{valentin.radu@sheffield.ac.uk}
\affiliation{%
  \institution{University of Sheffield}
  \country{}
}

\author{Po Yang}
\email{po.yang@sheffield.ac.uk}
\affiliation{%
  \institution{University of Sheffield}
  \country{}
}

\renewcommand{\shortauthors}{Hongrui Shi, et al.}
\renewcommand{\shorttitle}{Closing the Gap between Client and Global Model Performance in Heterogeneous FL}

\begin{abstract}
  
  The heterogeneity of hardware and data is a well-known and studied problem in the community of Federated Learning (FL) as running under heterogeneous settings. Recently, custom-size client models trained with Knowledge Distillation (KD) has emerged as a viable strategy for tackling the heterogeneity challenge. However, previous efforts in this direction are aimed at client model tuning rather than their impact onto the knowledge aggregation of the global model. Despite performance of global models being the primary objective of FL systems, under heterogeneous settings client models have received more attention. Here, we provide more insights into how the chosen approach for training custom client models has an impact on the global model, which is essential for any FL application. We show the global model can fully leverage the strength of KD with heterogeneous data. 
  Driven by empirical observations, we further propose a new approach that combines KD and Learning without Forgetting (LwoF) to produce improved personalised models. We bring heterogeneous FL on pair with the mighty FedAvg of homogeneous FL, in realistic deployment scenarios with dropping clients. 
  
\end{abstract}



\maketitle
\section{Introduction}\label{sec:introduction}

Federated Learning (FL) has become a popular machine learning paradigm for training models on many edge computing devices using locally generated data without compromising privacy. However, expanding FL to include millions of clients is a double-edge sword. While beneficial to model generalisation by experiencing diverse data distributions, working with many heterogeneous devices exposes the challenges of straggler clients and optimisation inconsistency~\citep{wang2020tackling, li2018federated}. Empirical studies show that extreme heterogeneous environments can degrade the performance of FL by up to 4$\times$~\citep{10.1145/3517207.3526969}. 

A major issue causing FL to underperform in heterogeneous settings is the distribution and aggregation of a single-architecture (structure) model across clients. To address this, emerging studies are focusing on distributing custom-size (personalised) models to each client. In~\citep{smith2017federated, liang2020think, jiang2019improving}, FL aggregates client updates of a fix model structure into a global model. Another solution is to distribute a custom-size models to each client, and train with knowledge distillation (KD), thus having heterogeneous models across the federated learning pool. This latter approach offers the opportunity to customise models on the computing characteristic of local hardware and data. KD-based FL has been explored in recent works, either for improving performance of FL with heterogeneity~\citep{li2019fedmd, lin2020ensemble, itahara2020distillation, chen2020fedbe, he2020group} or for achieving communication efficiency~\citep{jeong2018communication, guha2019one} and resilience against adversarial attacks~\citep{chang2019cronus, sun2020federated}, as well surveyed in~\citep{mcmahan2017communication, zhang2021parameterized}.

Our work focuses on training a server-side global model using an ensemble of knowledge distillations from personalised client models. The aforementioned research effort in the community either ignores the effect of client side choices onto the global model of the server~\citep{li2019fedmd, chang2019cronus} or relies on homogeneous model architectures for improving client performance~\citep{jeong2018communication, lin2020ensemble, chen2020fedbe}. Our expectation is that both server and client models should improve performance via KD-based ensemble training in heterogeneous settings. In our case, KD is performed on both server and client side to train the global model and client models respectively. The knowledge used by KD are the logits created by client models on a public dataset, aggregating on the server and distributing to clients every FL round. Following the KD training, custom client models are updated on the client side using the local data. 

In our evaluation, Cifar-100~\citep{Krizhevsky2009} is assigned as the public dataset to facilitate knowledge distillation, and fractions of Cifar-10~\citep{Krizhevsky2009} as local data on the client side. The two chosen datasets are disjoint to strengthen the data privacy characteristic of our solution. We simulate non-IID local data settings for the vanilla FedAvg~\citep{mcmahan2017communication} but considering a realistic scenario with computational impaired clients not participating in global updates. With our method, both global and client models have similar performance in global updates, with the global model even outperforming client models by 3\% to 4\%. In local updates with extreme non-IID settings, the performance of custom client models degrades substantially. To reduce the performance gap, we boost the training by integrating Learning without Forgetting (LwoF)~\citep{li2017learning} in our solution. LwoF has the effect of reducing sudden shifts in model representations, which results in improved client model performance by 5\%, and narrows the performance gap with the global model.

Our work making the following contributions:

\begin{itemize}
    \item We relax the condition of FL with a single architecture model distributed across clients by customising the architecture of each client model to match the computation resources of the target device. Our solution integrates KD to aggregate and transfer the knowledge from heterogeneous client models. 
    \item Our evaluation is designed to reflect realistic conditions of heterogeneous client hardware by assuming a part of clients dropping from FL training loop due to their insufficient computation resources.
    \item We bring the focus to global model performance during training, not just the client performance. We control the gap between global model performance and client model performance via LwoF, an effective solution to prevent models from forgetting previously learned knowledge.
\end{itemize}

\section{Related Work}

One of the major challenges FL faces is the heterogeneity of clients. In real world, both client hardware and data may vary significantly in terms of computation resources, network capacity and data distribution. This fact makes general FL working on hundreds or even millions of clients struggle to fit the local data characteristics and to meet fast synchronization time frames. Recently, a growing number of works has sought to approach heterogeneity via model personalisation, by tuning the global model in the local environment.

\subsection{Personalised Clients}
In contrast to using a universal model, personalised FL (pFL) adopts heterogeneous models that calibrate both model weights or architectures. The information exchanged in the network depends on the chosen underlying techniques.

Tasks in Multi-Task Learning (MTL) and Meta-Learning are comparable to the clients in FL. MOCHA ~\citep{smith2017federated} treats local updates of a client model as an individual learning task in the pipeline of MTL. It characterizes global updates by client parameter correlation, and local updates for personalising client models. The Model-Agnostic Meta-Learning (MAML)~\citep{finn2017model} is connected to pFL by comparing the samples of learning tasks from the meta-training phase to local updates in FL. Authors of~\citep{jiang2019improving} show that FL forms a good initial model, which can be further personalised via Meta-Learning. 

Split learning is also adopted to personalise models. In~\citep{liang2020think} the lower part of the global model is personalised locally, while the upper parts are updated using standard model fusion~\citep{mcmahan2017communication}. The personalised lower part is essentially a feature extractor adapted by each client. 

The aforementioned pFLs either fully or partially rely on a homogeneous model of a single size architecture. To broaden the application of FL to devices with smaller computing capability, research in custom size model for client side is just emerging in the community.

\subsection{Knowledge Distillation in FL}

One approach for allowing heterogeneous model architectures in FL is building on knowledge distillation (KD), also known as knowledge transfer (KT). 
As architectural differences prohibit the server from using direct model fusion to transfer the learned knowledge, KD exploits the inner knowledge of client models exposed by their outputs, without needing to align the architectures, just their outputs. 

Early applications of KD to FL ~\citep{jeong2018communication} are motivated by improving the communication efficiency. Compared to transmitting the whole model in general FL, KD-based FL features the transmission of logits, model outputs of far smaller size. Later efforts~\citep{chang2019cronus, sun2020federated} demonstrate the robustness of KD in adversary attacks and noise pollution, further solidifying its advantageous position. Even more, KD is helping to address the heterogeneity challenge due to its construction on personalised models. FedDF~\citep{lin2020ensemble} highlights architectural independence of client models by integrating KD. However, it still includes the model fusion before performing ensemble distillation, styling the model heterogeneity with architecture clusters. FedMD~\citep{li2019fedmd} and Cronus~\citep{chang2019cronus} produce good client models by relying on the knowledge transfer across heterogeneous models, but they do not showcase a global model for the server side, which is common to FL. At a different end, DS-FL~\citep{itahara2020distillation}, FedBE~\citep{chen2020fedbe} and KT-pFL~\citep{zhang2021parameterized} contribute advanced methods to aggregating the logits in KD-based FL, mitigating the problem of data heterogeneity.




\section{Methods}\label{sec:methods}
\subsection{System Setup}
Our objective is to train a global model on the server using knowledge distilled from personalised client models of different architectures. Algorithm~\ref{alg: our method} presents the proposed solution. All client models $M_{k}, \; k = 1\dots N$ and the global model $M_{g}$ have the same output size (to apply KD), but with varying lower architectures. The learning task is defined by the local data $\mathcal{D}_{k} ,\; k = 1\dots N$ accommodated by clients. Knowledge distillation is used to transfer the global knowledge to individual models. To facilitate the knowledge transfer, we follow prior approaches~\citep{li2019fedmd, chen2020fedbe, lin2020ensemble}, employing a public dataset $\mathcal{D}_{p}$ and indiscriminately distributing it to the server and clients. We organise our solution into two stages, the global update and the local update, introduced in Section ~\ref{sec:GU} and Section ~\ref{sec:LU} respectively.

\RestyleAlgo{ruled}
\SetKwComment{Comment}{/* }{ */}
\begin{algorithm}
\caption{FedMD+Global}
\textbf{Input:} Public data $\mathcal{D}_{p}$, Local data $\mathcal{D}_{k}$, personalised client model $M_{k}, \; k=1 \ldots N$\;
\textbf{Initialization:} Train $M_{k},\; k=1 \ldots N$ on $\mathcal{D}_{p}$ and then on $\mathcal{D}_{k}$ until converge (a pre-FL stage detailed in Section~\ref{sec:experiments})\;
\For{FL round $t=1, 2, ...$}{
\textbf{Global Updates:}

    \For{Client $M_{k},\; k=1 \ldots N$}{
    $\bm{z}^{p}_{k} \leftarrow$ \textbf{Logits} $\left(M_{k}, \: \mathcal{D}_{p}\right)$\;
    }
Server collects $\bm{z}^{p}_{k}, \; k = 1\dots N$, create the global knowledge $\hat{\bm{z}}^{p} = \frac{1}{N} \sum_{k} \bm{z}^{p}_{k}$\;
$M^{\prime}_{g} \leftarrow$ \textbf{Train} $\left(M_{g}, \: \hat{\bm{z}}^{p}, \: \bm{z}^{p}_{g}\right)$ with Eq. ~\ref{Eqa. GlobalUpdates} as loss\;
\For{ Client $M_{k},\; k=1 \ldots N$}{
    Downloads $\hat{\bm{z}}^{p}$ from server\;
    $M^{\prime}_{k} \leftarrow$ \textbf{Train} $\left(M_{k}, \: \hat{\bm{z}}^{p}, \: \bm{z}^{p}_{k}\right)$ with Eq. ~\ref{Eqa. GlobalUpdates} as loss\;
    }

\textbf{Local Updates:}

    \For{$M_{k},\; k=1 \ldots N$}{
    $M_{k} \leftarrow$ \textbf{Train} $\left(M^{\prime}_{k}, \:  \mathcal{D}_{k}\right)$ with Eq. ~\ref{Eqa. LocalUpdates} or Eq. ~\ref{Eqa. LocalUpdates with LwF} if LwoF is applied\;
    }
}
Return $M_{k}, k=1 \ldots m$
\label{alg: our method}
\end{algorithm}

\subsection{Global Updates}\label{sec:GU}
Global updates perform ensemble knowledge distillation for global model and client models, gathering and transferring the global knowledge. The global knowledge is in the form of logits distilled by the client models on the public data, $\mathcal{D}_{p}$. Every client model $k \in \{1, 2, ..., N\}$ computes a forward pass on the instances of $\mathcal{D}_{p}$ to generate its logits, $\bm{z}^{p}_{k}$, model outputs ahead of the top softmax layer. The server collects $\bm{z}^{p}_{k},\; k = 1\dots N$ from clients and averages them, $\hat{\bm{z}}^{p} = \frac{1}{N} \sum_{k} \bm{z}^{p}_{k}$. $\hat{\bm{z}}^{p}$ is used as the global knowledge to update $M_{g}$ and $M_{k}, \; k = 1\dots N$ in parallel using a loss function as follows. 

\begin{equation}
\mathcal{L}_{g}\left(\hat{\bm{z}}^{p}, \bm{z}^{p}_{s}\right) = \|\hat{\bm{z}}^{p}-\bm{z}^{p}_{s}\|_{1}
\label{Eqa. GlobalUpdates}
\end{equation}



where $\mathcal{L}_{g}$ \footnote{Our work employs $\mathcal{L}_{1}(\cdot)$ loss function. However, other loss functions such as $\mathcal{L}_{C E}(\cdot)$ and $\mathcal{L}_{2}(\cdot)$ can also be applied.} is the loss function for model training, $\bm{z}^{p}_{s} \in \{\bm{z}^{p}_{1}, \bm{z}^{p}_{2}, ..., \bm{z}^{p}_{N}, \bm{z}^{p}_{g}\}$ \footnote{The notation is abused for simplicity. Client logits in training are different to previous ones used for aggregation because the model is being updated.} are the logits from client models and the global model. Training with the loss function, the model is optimised to minimise the differences of its own logits $\bm{z}^{p}_{s}$ and the aggregated logits $\hat{\bm{z}}^{p}$ on $\mathcal{D}_{p}$. 

The global updates do not break the privacy-persevering principle of FL because they only rely on $\mathcal{D}_{p}$ and communicate the logits. By approaching $\hat{\bm{z}}^{p}$, $M_{g}$ and $M_{k}, \; k = 1\dots N$ learn the distilled global knowledge without visiting local data. The performance at this stage will be evaluated in Section \ref{sec:experiments}.

\subsection{Local Updates and LwoF}\label{sec:LU}
The global updates are followed by local updates that optimise client model $M_{k}$ on its associated local data $\mathcal{D}_{k}$, personalising the model performance as follows.

\begin{equation}
    \argmin{\bm{W}_{k}} F (\bm{W}_{k}, \: \mathcal{D}_{k}) = \mathcal{L}_{t}(f_{k}(\bm{W}_{k}, \:\bm{x}_k),\: \bm{y}_k) 
\label{Eqa. LocalUpdates}
\end{equation}

where $\mathcal{L}_{t}$ is the loss function for the target task. $\bm{W}_{k}$ and $f_{k}$ are the weights and function represented by client model $M_{k}$. $(\bm{x}_k, \bm{y}_k)$ is the instance of input-target pairs of local data $\mathcal{D}_{k}$. 

The learned global knowledge could be lost in local updates. To protect it, we add a regularisation term, shown in Equation ~\ref{Eqa. LwF term}, which is commonly known as Learning without Forgetting (LwoF)~\citep{li2017learning}, to Equation ~\ref{Eqa. LocalUpdates}. On instances of $\mathcal{D}_{k}$, the regularisation term calculates the cross entropy of the soft labels created by a copy of $M_{k}$ in the previous global update stage and the $M_{k}$ currently being updated.

\begin{equation}
\mathcal{L}_{C E} (\bm{x}_k ; \rho)=- p(\bm{x}_k; \rho) \log (q(\bm{x}_k; \rho))
\label{Eqa. LwF term}
\end{equation}

where $p(\bm{x}_k;\rho)=\frac{\exp \left(\frac{\bm{z}^{\prime}_{k}}{\rho}\right)}{\sum_{j} \exp \left(\frac{\bm{z}^{\prime}_{k}}{\rho}\right)}$ and $q(\bm{x}_k; \rho)=\frac{\exp \left(\frac{\bm{z}_{k}}{\rho}\right)}{\sum_{j} \exp \left(\frac{\bm{z}_{k}}{\rho}\right)}$, $\bm{z}_{k}$ stands for the logits of $M_{k}$, $\bm{z}^{\prime}_{k}$ represents the logits calculated by the $M_{k}$ copied from the global update stage. $\rho$ is the temperature, a hyperparameter used to further soften the outputs of the softmax. With the additional term, the objective function of local updates, Equation~\ref{Eqa. LocalUpdates}, becomes

\begin{equation}
    \argmin{\bm{W}_{k}} F (\bm{W}_{k}, \: \mathcal{D}_{k}) = \mathcal{L}_{t}(f_{k}(\bm{W}_{k}, \:\bm{x}_k), \bm{y}_k) + \beta \mathcal{L}_{C E} (\bm{x}_k ; \rho) 
\label{Eqa. LocalUpdates with LwF}
\end{equation}

The regularisation term added to the local update prevents $M_{k}$ from deviating significantly from the $M_{k}$ updated by the global knowledge $\hat{\bm{z}}^{p}$, protecting the global knowledge. $\beta$ is the coefficient for the regularisation term. In our evaluation, it is set to 1 as originally introduced in~\citep{li2017learning}. 
\section{Experiments}\label{sec:experiments}

\textbf{Task and Data.} We evaluate our method on an image classification task with Cifar-10 and Cifar-100~\citep{Krizhevsky2009} as the local (client) data and public (server) data respectively. Cifar-10 has a total number of 60,000 images of size $32\times32$ pixels, split across 10 classes with 6,000 images each. Its training set consists of 50,000 images and the test set contains 10,000 images. Cifar-100 is a harder dataset than Cifar-10 in the number of classes, containing 100 classes, each with just 600 samples. For simplicity, we use the coarse labels of Cifar-100 instead of its fine labels, by grouping the 100 fine classes into 20 coarse classes, each group containing 5 fine classes under a single label. Throughout this section, the performance of models is evaluated on the test set of Cifar-10. 

\textbf{Baselines.} We compare our method with FedAvg~\cite{li2019convergence}, one of the most popular FL baselines in previous studies of heterogeneous FL. We design the evaluation to consider the impact of clients running on small devices with constrained computing resources. For our method, we customise the client model size such that its participation in the training loop is guaranteed by fitting the available computing resources of the client. In contrast, FedAvg distributes a single-size model to all its clients. To simulate smaller clients struggling to perform the required computations on the single size model of FedAvg, we introduce client dropping from the training round. This is equivalent to some clients not meeting the round deadline before the global aggregation. We experiment with three FedAvg client participation settings: 100\%, 60\%, 40\%. Considering that vanilla FedAvg~\citep{mcmahan2017communication} was designed to operate with as low as 10\% client participation, our baseline is not impairing FedAvg.

\textbf{Heterogeneous Models.} All our models are constructed on the Wide Residual Networks (WRN)~\cite{zagoruyko2016wide} with varying depth and width. For the global (server) model, we choose a WRN-40-1 with depth of 40 (layers) and width 1 (full). For the 20 custom-size client models we vary the depth of WRN between 10 and 40, as presented in Table~\ref{tab:Model Stat}. Clients with less computation resources receive a shallower WRN, while the largest client model is a WRN-34-1 for the more capable computing clients. In the case of FedAvg evaluation, all the client models share the same structure as the global model, a WRN-40-1. By using heterogeneous models, our method enjoys a smaller overall memory footprint compared with the baseline FedAvg of 60\% and 100\% client participation.

\begin{figure*}[tb]

\begin{subfigure}{.4\textwidth}
\centering
\centerline{\includegraphics[width=1\columnwidth]{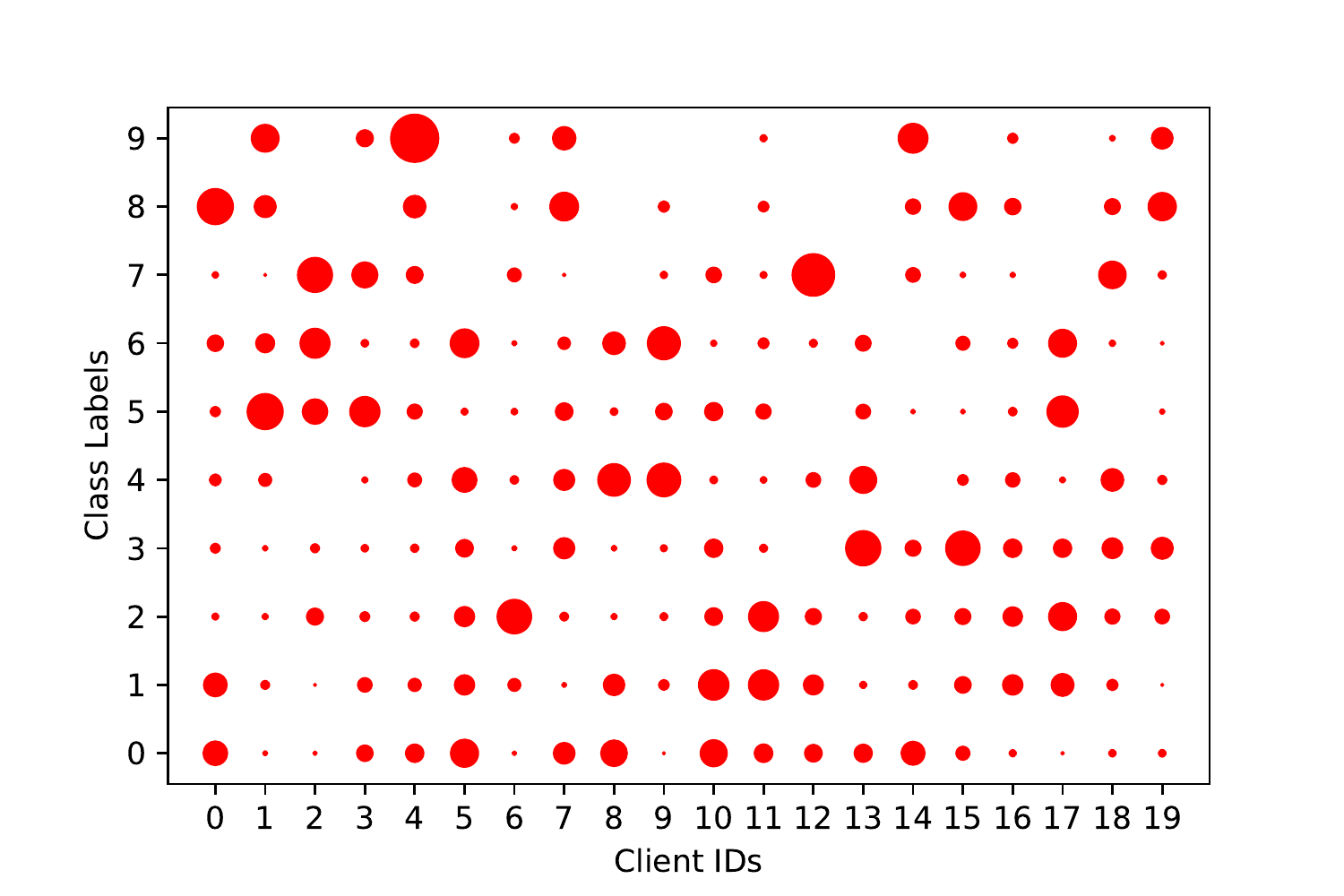}}
\caption{non-IID distribution with $\alpha = 0.5$.}
\label{fig:non-IID alpha=0.5}
\end{subfigure}
\begin{subfigure}{.4\textwidth}
\centering
\centerline{\includegraphics[width=1\columnwidth]{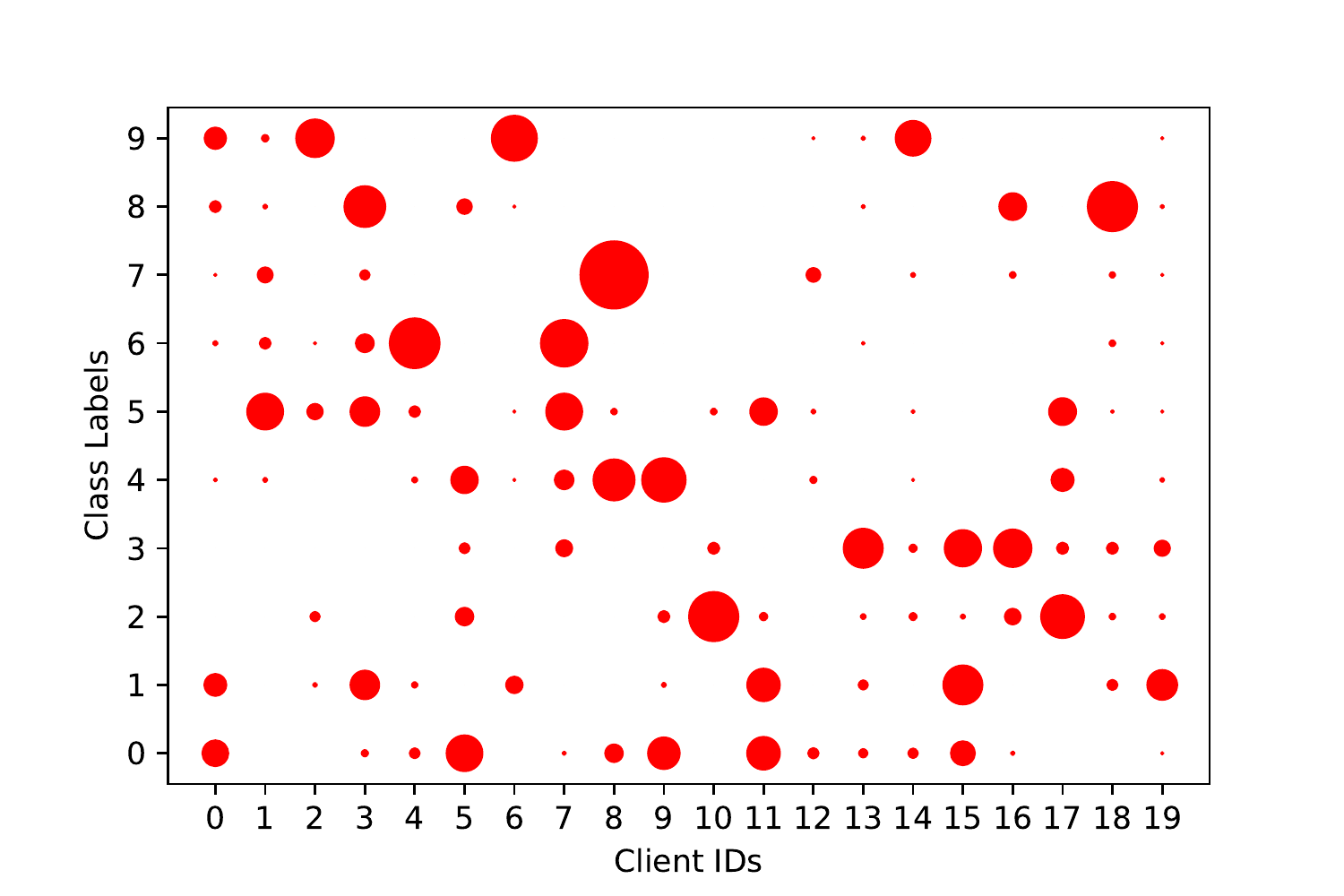}}
\caption{non-IID distribution with $\alpha = 0.1$.}
\label{fig:non-IID alpha=0.1}
\end{subfigure}

\caption{The visualization of non-IID local data across 20 clients. Client ID are presented on the Ox axis, and on Oy the Cifar-10 classes. The size of the bubble indicates the number of samples. For the higher $\alpha=0.5$, data is approaching an IID case. In contrast, $\alpha=0.1$ introduces a high degree of non-IIDness, shown by the varying sizes of bubbles.}
\label{fig:non-IID visualization}

\end{figure*}

\textbf{Heterogeneous Data.} Our evaluation adopts the Dirichlet distribution in splitting the non-IID local data, following the path set by previous studies~\citep{lin2020ensemble, he2020group, hsu2019measuring}. The hyperparameter $\alpha$ of the distribution embodies the level of non-IIDness. The larger $\alpha$ is, the more uniformly distributed classes are in client local data. Figure~\ref{fig:non-IID visualization} illustrates the two data distribution scenarios, $\alpha=0.5$ and $\alpha=0.1$, with splits of data across the 20 clients. A larger bubble indicates a larger number of data samples of the same class present in the local data of a client ID. We chose these two scenarios to show the impact of data heterogeneity.



\begin{table}[]
\begin{center}
\caption{Model configuration of our method and for the Fedavg baselines with 20 clients. Our method features WRNs with varying depths, custom to the client computing resources, significantly reducing the total number of trainable parameters.}
\begin{tabular}{@{}cc|ccc@{}}
\toprule
Method& Parties  & Depth &Width & \#Para\\ \midrule
Ours &\begin{tabular}{@{}c@{}}Client 0-5 \\ Client 6-9\\Client 10-13 \\Client 14-17\\Client 18, 19\\ Global\\Total\end{tabular}&\begin{tabular}{@{}c@{}}10 \\ 16\\22\\28\\34\\40\\-\end{tabular} & \begin{tabular}{@{}c@{}}1 \\ 1\\1\\1\\1\\1\\-\end{tabular} & \begin{tabular}{@{}c@{}}79k \\ 177k\\274k\\372k\\469k\\567k\\\textbf{4.7m}\end{tabular} \\ \midrule
FedAvg& \begin{tabular}{@{}c@{}}Global\&Client\\40\%/60\%/100\% Total\end{tabular}& \begin{tabular}{@{}c@{}}40\\-\end{tabular} & \begin{tabular}{@{}c@{}}1\\-\end{tabular}& \begin{tabular}{@{}c@{}}567k\\3.8/5.6/9.4m\end{tabular} \\\midrule
\end{tabular}

\label{tab:Model Stat}
\end{center}
\end{table}




\textbf{Optimisation Setup.} Our experiments apply a uniform optimisation setup in global updates, with varying conditions for the local update hyperparameters. A round of global FL update include 5 epochs for knowledge distillation. To achieve a faster global updates, only 5000 data points are randomly sampled from the public data (10\% of the Cifar-100 training set). For the local updates, we experiment with 5 and 10 as the number of local epochs, for both levels of heterogeneity $\alpha = 0.5/0.1$. SGD with a learning rate of 0.1 is employed as the optimiser for both global and local updates. The batch sizes for global and local updates are tuned to 64 and 128 respectively. 

\textbf{Model Initialisation.} To accelerate the system convergence, we follow the practice introduced by~\citep{li2019fedmd}, initializing client models before the commencement of FL. Every client model is trained on the public data and then on its local data. We define as \emph{Initial (performance)} the performance of the client model after transfer from server from global model. For initialization, the size of the WRN output is set to 30, equal to the sum between the number of classes in Cifar-100 (20) and the number of classes in Cifar-10 (10). This multi-task approach joins the use of two disjoint datasets. The initial performance is not influenced by the number of local update epochs. For the FedAvg baseline, the initialisation involves only the local data as FedAvg does not use a public dataset.

\begin{figure*}[tb]
\vskip 5mm
\centerline{\includegraphics[width=1.6\columnwidth]{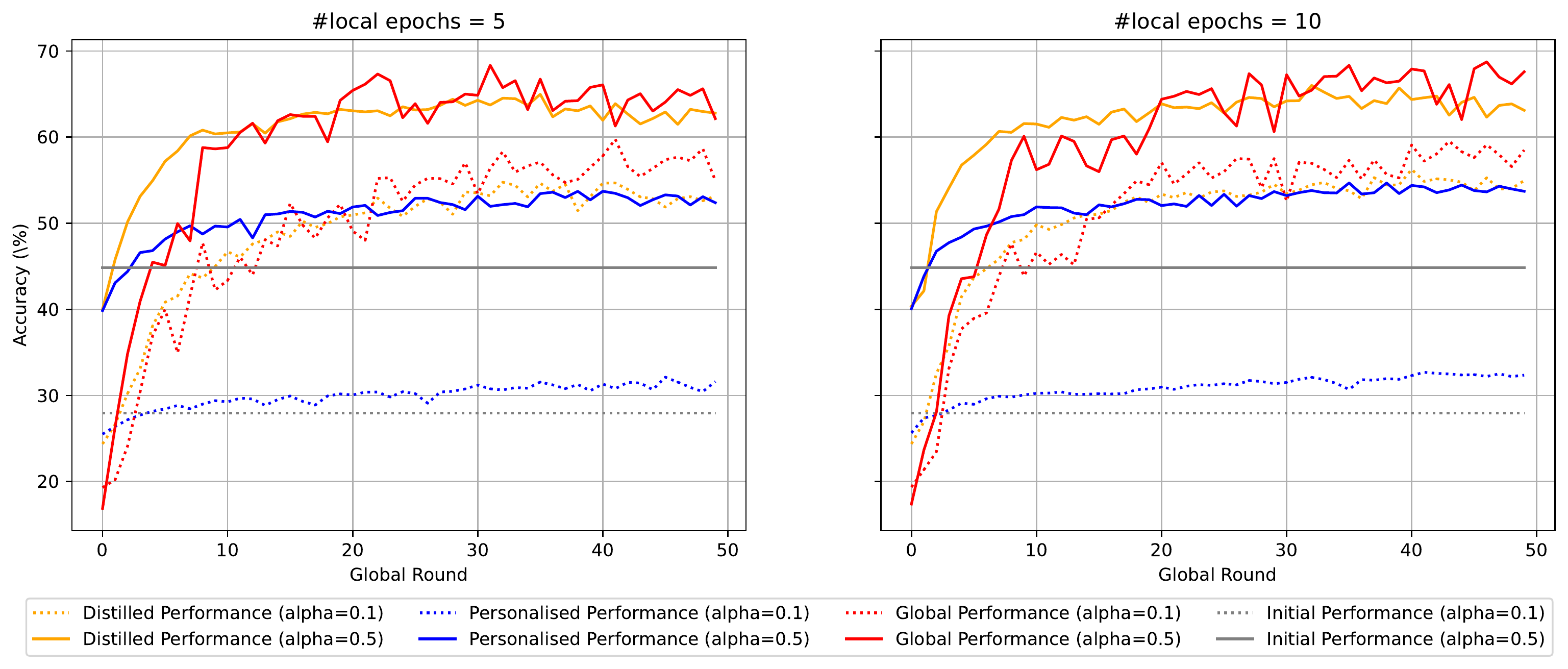}}

\caption{Visualizing the performance of the global model and the average across the 20 client models on the Cifar-10 test data, as detailed in Table~\ref{tab:fedmd with global}.}
\label{fig:fedmd with global}
\vskip -5mm
\end{figure*}

\textbf{Experiment Results.} We run all experiments for 50 global rounds. Table~\ref{tab:fedmd with global} and Figure~\ref{fig:fedmd with global} present the experiment results for our method introduced in the previous sections and that of the FedAvg baseline. 
The performance of the global model is presented as \emph{global performance}, while the performance of client models is presented as the average performance at the stage of local updates and global updates with distillation, as \emph{personalised performance} and \emph{distilled performance} respectively. 
Given $\alpha=0.1$, the global model in our method outperforms the baseline with 40\% client participation by a large margin (7.28\%) and similar performance to the baseline with 60\% client participation, just 0.7\% lower. The huge client performance increase from initialisation to distillation, more than 31\% ($\alpha=0.1$) and 23\% ($\alpha=0.5$), denote the effectiveness of KD in transferring the global knowledge. We also see the global model trained by KD outperforms client models at global updates stage, by 3\% to 4\%. These results evidence the feasibility of training a global model without relying on single-size model aggregation as commonly done in previous works. Finally, the number of local epochs does not affect the global model performance greatly, except for the client models, which enjoys a minor performance increase (around 2\%) for $le=10$. 

\textbf{The Impact of Data Heterogeneity.} The performance of the distilled model is constantly better than the performance of the personalised model, while a high degree of data heterogeneity leads to a larger gap between them. 
The gap jumps from 11.26\% to 22.64\%. This indicates that the client model is better pruned to forget the learned global knowledge during its training on local data with heterogeneity. Additionally, when given a mild data heterogeneity ($\alpha=0.5$), we find our solution is outperformed by the baseline with 40\% client participation. From this, we observe that in low data heterogeneity the performance of FedAvg is not much affected by clients dropping, but if data heterogeneity is high, our method works better. 

\begin{table}[]
\begin{center}
\caption{The performance of the global model and that of client models on the Cifar-10 test data. \emph{Global performance} is the test accuracy achieved by the global model at the distillation stage. \emph{Distilled performance} and \emph{personalisation performance} are the average test accuracy of the 20 client models at the global update stage and local update stage, respectively. \emph{Initial performance} is the average test accuracy of client models from initialisation. Baseline performance is the test accuracy achieved by the global model of FedAvg with 100\%/60\%40\% client participation. Bold values for our method indicate the best performance achieved by the global model with our method. FedAvg (60\%) has some values close to our method, while FedAvg (100\%) is an unrealistic scenario to have no client drops.}
\vskip -2mm
\begin{tabular}{@{}c|cc@{}}
\toprule
Methods &\begin{tabular}[c]{@{}c@{}}$le=5$\\ \midrule \begin{tabular}{@{}cc@{}}$\alpha=0.5$&$\alpha=0.1$ \end{tabular} \end{tabular} &\begin{tabular}[c]{@{}c@{}}$le=10$\\ \midrule \begin{tabular}{@{}cc@{}}$\alpha=0.5$&$\alpha=0.1$ \end{tabular} \end{tabular} \\ \midrule
FedAvg \;(100\%)  & \begin{tabular}{@{}cc@{}}79.29\% & 65.59\% \end{tabular} & \begin{tabular}{@{}cc@{}}79.15\% & 66.84\% \end{tabular}  \\ 
FedAvg \;(60\%)  & \begin{tabular}{@{}cc@{}}76.07\% & \textbf{60.4}\% \end{tabular} & \begin{tabular}{@{}cc@{}}76.00\% & \textbf{60.29}\% \end{tabular}  \\ 
FedAvg \;(40\%)  & \begin{tabular}{@{}cc@{}}72.20\% & 52.48\% \end{tabular} & \begin{tabular}{@{}cc@{}}71.97\% & 49.16\% \end{tabular} \\ \midrule
Global \; (ours)  & \begin{tabular}{@{}cc@{}}\textbf{68.34}\% & \textbf{59.76}\% \end{tabular} & \begin{tabular}{@{}cc@{}}\textbf{68.74}\% & \textbf{59.53}\% \end{tabular}  \\
Initial\; (ours)  & \begin{tabular}{@{}cc@{}}44.87\% & 27.95\% \end{tabular} & \begin{tabular}{@{}cc@{}}44.87\% & 27.95\% \end{tabular} \\
Distilled \; (ours)  & \begin{tabular}{@{}cc@{}}64.98\% & 54.77\% \end{tabular} & \begin{tabular}{@{}cc@{}}66.01\% & 56.27\% \end{tabular}  \\ 
Personalised \; (ours)  & \begin{tabular}{@{}cc@{}}53.72\% & 32.13\% \end{tabular} & \begin{tabular}{@{}cc@{}}54.68\% & 32.69\% \end{tabular}  \\ 
Gap (Dis. \& Per.)  & \begin{tabular}{@{}cc@{}}11.26\% & 22.64\% \end{tabular} & \begin{tabular}{@{}cc@{}}11.33\% & 23.58\% \end{tabular}  \\ \midrule

\end{tabular}
\vskip -5mm

\label{tab:fedmd with global}
\end{center}
\end{table}

\begin{figure*}[tb]
\vskip 5mm
\centerline{\includegraphics[width=1.6\columnwidth]{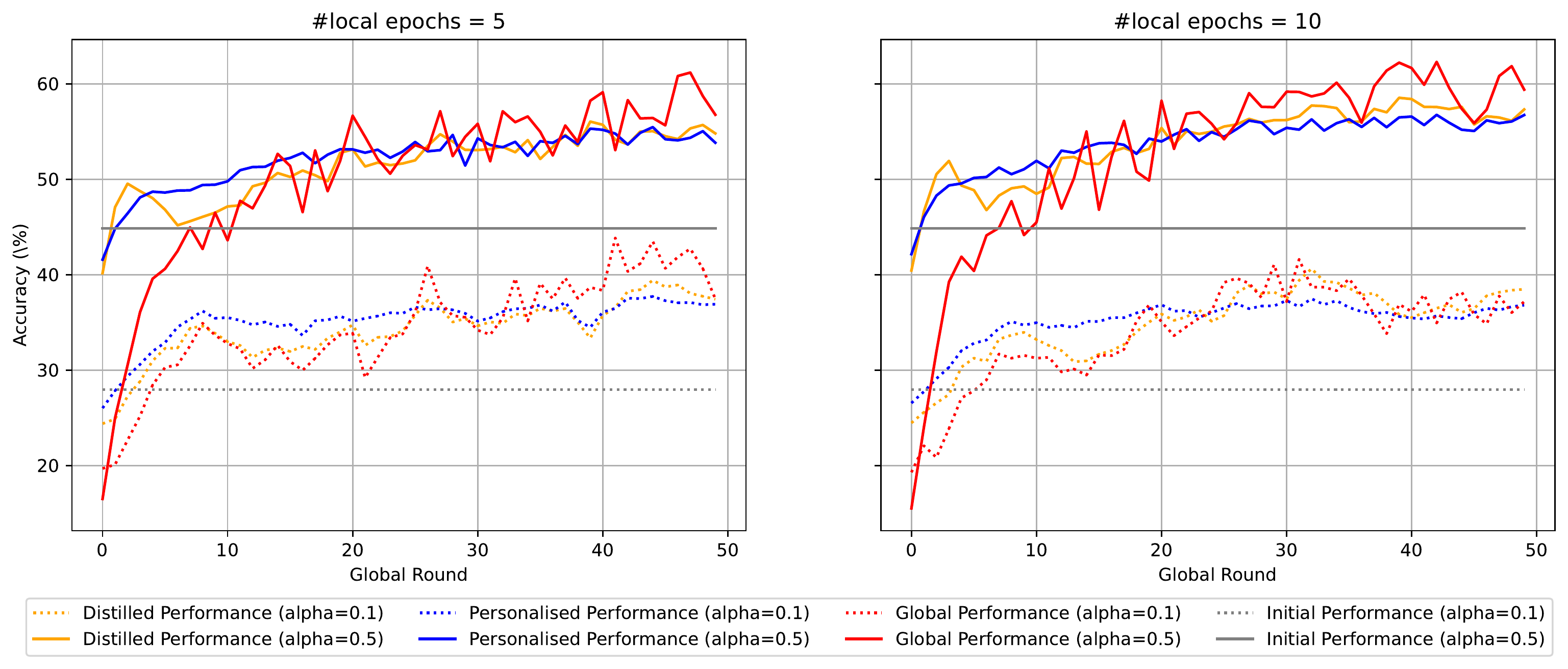}}

\caption{Visualising the performance of the global model and the average across the 20 clients when applying the LwoF to local updates, as detailed in Table~\ref{tab:fedmd with global and lwf}.}
\label{fig:fedmd with global and lwf}
\vskip -5mm
\end{figure*}

\textbf{Application of LwoF.} We further adopt LwoF in local updates to address the forgetting problem of global knowledge. Table ~\ref{tab:fedmd with global and lwf} and Figure ~\ref{fig:fedmd with global and lwf} summarise the experiment results. Here our method uses equation~\ref{Eqa. LocalUpdates with LwF}. First, the performance gap between distillation and personalisation is largely reduced, dropping to less than 3.15\% from over 23\% in all scenarios, proving the effectiveness of LwoF. 
Second, the performance of personalised client model is substantially improved, particularly given a high degree of local data heterogeneity, 5\% improvement for $\alpha=0.1$. 
However, narrowing the performance gap sacrifices the global and distilled performance, decreasing by up to 17\%. As such, LwoF is applicable in our proposed method if the client values the performance of the personalised model. Otherwise, the client model learned at the distillation stage is more favorable. 

\begin{table}[]
\begin{center}
\vspace{0.5cm}
\caption{The performance of the global model and of the 20 client models on Cifar-10 test data when \textbf{applying LwoF}. Bold figures present the best performance achieved by the global model with our method. The arrows indicate the change of performance compared with Table~\ref{tab:fedmd with global}. }
\begin{tabular}{@{}c|cc@{}}
\toprule
Performance &\begin{tabular}[c]{@{}c@{}}$le=5$\\ \midrule \begin{tabular}{@{}cc@{}}$\alpha=0.5$&$\alpha=0.1$ \end{tabular} \end{tabular} &\begin{tabular}[c]{@{}c@{}}$le=10$\\ \midrule \begin{tabular}{@{}cc@{}}$\alpha=0.5$&$\alpha=0.1$ \end{tabular} \end{tabular} \\ \midrule
Initial  & \begin{tabular}{@{}cc@{}}44.87\% & 27.95\% \end{tabular} & \begin{tabular}{@{}cc@{}}44.87\% & 27.95\% \end{tabular}  \\ \midrule
Global  & \begin{tabular}{@{}cc@{}}\textbf{61.18}\% $\downarrow$& \textbf{43.84}\% $\downarrow$\end{tabular} & \begin{tabular}{@{}cc@{}}\textbf{62.3}\% $\downarrow$& \textbf{41.60}\% $\downarrow$\end{tabular}  \\
Distilled  & \begin{tabular}{@{}cc@{}}56.04\% $\downarrow$& 39.41\% $\downarrow$\end{tabular} & \begin{tabular}{@{}cc@{}}58.54\% $\downarrow$& 40.62\% $\downarrow$\end{tabular}  \\ 
Personalised  & \begin{tabular}{@{}cc@{}}55.46\% $\uparrow$& 37.71\% $\uparrow$\end{tabular} & \begin{tabular}{@{}cc@{}}56.75\% $\uparrow$& 37.47\% $\uparrow$\end{tabular}  \\ \midrule
Gap (Dis. \& Per.)  & \begin{tabular}{@{}cc@{}}0.58\% $\downarrow$& 1.70\% $\downarrow$\end{tabular} & \begin{tabular}{@{}cc@{}}1.79\% $\downarrow$& 3.15\% $\downarrow$\end{tabular}  \\ \midrule

\end{tabular}

\label{tab:fedmd with global and lwf}
\end{center}
\end{table}

\section{Conclusion}
In this paper, we focus on the training of a global model in heterogeneous client settings, driven by knowledge distillation. The performance is evaluated at the stages both distillation on the server and in local updates on the client. Using non-IID local data from Cifar-10, and Cifar-100 for knowledge distillation on the server, the global model achieves comparable performance to that of FedAvg, which benefits from a high client participation of 60\%. Our experiments also highlight the susceptibility of client models being affected by the forgetting in extreme non-IID settings. We show that the performance of client models can be immediately improved with the adoption of LwoF, though in detriment to global model performance. 

\balance
\bibliographystyle{ACM-Reference-Format}
\bibliography{sample-base}










\end{document}